\documentclass[lettersize,journal]{IEEEtran}
\usepackage{amsmath,amsfonts}
\usepackage{algorithmic}
\usepackage{algorithm}
\usepackage{array}
\usepackage[caption=false,font=normalsize,labelfont=sf,textfont=sf]{subfig}
\usepackage{textcomp}
\usepackage{stfloats}
\usepackage{url}
\usepackage{verbatim}
\usepackage{graphicx}
\usepackage{cite}
\usepackage{hyperref}

\hypersetup{
    colorlinks=true,
    linkcolor=blue,
    filecolor=magenta,      
    urlcolor=cyan,
    citecolor=blue, 
    pdfstartview=FitV,
    pdftitle={My Title},
    pdfauthor={Author},
    pdfsubject={Subject},
    pdfkeywords={keyword1, keyword2, keyword3},
    pdfpagemode=None,
    pdfnewwindow=true 
}
\usepackage{amsmath,amsfonts}
\usepackage{algorithmic}
\usepackage{algorithm}
\usepackage{array}
\usepackage[caption=false,font=normalsize,labelfont=sf,textfont=sf]{subfig}
\usepackage{textcomp}
\usepackage{stfloats}
\usepackage{url}
\usepackage{verbatim}
\usepackage{graphicx}
\usepackage{cite}
\usepackage{booktabs}
\usepackage{color}
\usepackage{makecell}
\begin{document}

\title{Image-Conditional Diffusion Transformer for\\Underwater Image Enhancement}

\author{Xingyang Nie, Caoliang Zhang, Xiaoyu Zhai, Fengzhong Qu, Biao Wang, and Huilin Ge
\thanks{Xingyang Nie, Caoliang Zhang, Biao Wang, and Huilin Ge, are with the Ocean College, Jiangsu University of Science and Technology, Zhenjiang 212003, China. E-mail: 122386231@qq.com; 241112201127@stu.just.edu.cn; {wbiaoang,gh1989}@just.edu.cn .}

\thanks{Xiaoyu Zhai is with Nanjing Research Institute of Electronic Equipment, China Aerospace Science and Industry Corporation, Nanjing 210000, China. E-mail: 573911383@qq.com.}
\thanks{Fengzhong Qu is with the Ocean College, Zhejiang University, Zhoushan 316021, China E-mail: jimqufz@zju.edu.cn.}
\thanks{(Corresponding author: Xingyang Nie).}}



\maketitle

\begin{abstract}
Underwater image enhancement (UIE) has attracted much attention owing to its importance for underwater operation and marine engineering.
Motivated by the recent advance in generative models, we propose a novel UIE method based on image-conditional diffusion transformer (ICDT). Our method takes the degraded underwater image as the conditional input and converts it into latent space where ICDT is applied.
ICDT replaces the conventional U-Net backbone in a denoising diffusion probabilistic model (DDPM) with a transformer, and thus inherits favorable properties such as scalability from transformers.
Furthermore, we train ICDT with a hybrid loss function involving variances to achieve better log-likelihoods, which meanwhile significantly accelerates the sampling process.
We experimentally assess the scalability of ICDTs and compare with prior works in UIE on the Underwater ImageNet dataset.
Besides good scaling properties, our largest model, ICDT-XL/2, outperforms all comparison methods, achieving state-of-the-art (SOTA) quality of image enhancement.
\end{abstract}




\section{Introduction}
\label{sec1}
\IEEEPARstart
{U}{NDERWATER}  imaging has been extensively used in underwater archaeology, marine detection, and underwater robotics\cite{ref1,ref2,ref3}. However, distance- and wavelength-dependent light scattering and attenuation cause the problem of low contrast and color deviation in underwater images. These degraded images not only lead to unsatisfactory visual experience for humans but also affect the performance in computer vision tasks like semantic segmentation, object detection, and image classification.

To enhance the quality of underwater images, researchers have proposed a series of effective measures, promoting the development of underwater image enhancement (UIE). Conventional UIE methods (e.g., white balance\cite{ref4,ref5} and histogram equalization\cite{ref6}) depend on assumptions or a priori knowledge, models, or design guidelines to enhance underwater images. Although conventional methods are easy to execute, they are not adaptable to different lighting conditions and water environments.

The rapid development of deep learning brings new tools and insights for UIE\cite{iotj1}. Through exploiting large-scale underwater image datasets to learn the patterns and features, deep-learning-based methods for UIE can realize end-to-end, intelligent, and automated underwater image enhancement. Deep-learning-based UIE methods are typically divided into two primary categories: 1) convolutional neural network (CNN)-based and 2) generative adversarial network (GAN)-based. The CNN-based UIE methods train deep CNNs to learn the mapping relationship from the degraded image to the high-quality reference image\cite{ref7}, which is robust to different underwater scenes. The GAN-based UIE methods can also accomplish the conversion from the degraded underwater image to the corresponding ground truth image and achieve good performance\cite{ref8}. However, GANs often suffer from unstable training process and mode collapse\cite{iotj2, iotj3}. Therefore, it is necessary to develop more stable and diversified methods to further improve the effect and quality of UIE.

Denoising diffusion probabilistic models (DDPM) have begun to attract extensive attention for its good convergence properties and relatively stable training process. DDPMs have become a new state-of-the-art (SOTA) baseline in the field of image generation\cite{ref9,ref10}. As scholars continue to explore diffusion models, their potential for image-to-image generation is continually developed\cite{ref11}. In image restoration, image coloring, and image super-resolution tasks, diffusion model-based approaches have produced superior results than GAN models\cite{ref9,ref10,ref11}. Although DDPMs can yield excellent results, an acknowledged challenge of DDPMs is their poor real-time performance due to the long inference time.

In this paper, a generative approach for UIE based on conditional DDPM is proposed, which uses the degraded image as the conditional input. As for the model architecture, the conventional U-Net\cite{ref12} backbone in a diffusion model is replaced with a transformer, which we call image-conditional diffusion transformer (ICDT). Diffusion transformer adheres to many good practices of vision transformers (ViT)\cite{ref13}, one of the most important factors would be the excellent scalability properties compared to traditional convolutional networks.
Moreover, we incorporate learnt variances into DDPMs with reference to\cite{ref14}, allowing DDPMs to sample in much fewer steps without sacrificing sample quality.

The main contributions of this article are summarized as follows:
\begin{enumerate}
    \item We use a transformer as the backbone of the DDPM instead of the U-Net, which retains favorable properties like scalability from ViTs. The scaling performance is explored through experiments. We find that the sample quality is highly correlated with model complexity measured by FLOPs. Images of different quality generated by models with different complexity can meet the requirements for various downstream tasks (e.g., object detection, image classification, and semantic segmentation) and scenes, achieving a reasonable match between image-quality requirement and computing power.
    Furthermore, our most computational intensive model achieves a SOTA result on the standard benchmark dataset (Underwater ImageNet).
    \item The degraded image is utilized as an extra conditional input to the DDPM for UIE. It is converted to latent via a variational autoencoder (VAE) along with the corresponding reference image. The generated conditional latent and noised latent are then concatenated channel-wise. The transformer is applied to the latent space, constituting a computational-efficient hybrid architecture referred to as ICDT.
    The proposed ICDT has the potential to become a universal framework for image-to-image generation tasks like image translation, image coloring, image super-resolution, image restoration, and image inpainting.
    \item A hybrid loss function which provides learning signal for variances is employed in training of ICDT to achieve better log-likelihoods. Furthermore, incorporating learnt variances into DDPMs can considerably speed up sampling with a negligible change in sample quality, which is important for deployment in practical applications.
\end{enumerate}

This paper is organized as follows. In Section II, the related work is reviewed. Section III describes the proposed ICDT in detail. In Section IV, the experimental setup is presented.
Extensive experiments are performed to explore the scaling properties and demonstrate the performance of ICDT for UIE in Section V. Section VI concludes the paper.

\section{Related Work}
\subsection{Conventional UIE Method}
Conventional UIE methods depend on models, design rules, assumptions, or a priori knowledge to enhance underwater images. These methods are based on physical models and utilize degradation priors to inversely solve degradation models.
Drews et al.\cite{ref15} introduced a UIE method using underwater dark channel prior (DCP), which utilizes the statistical prior of images obtained in outdoor natural scenes and applies it only in the blue and green channels to improve enhancement quality.
Li et al.\cite{ref6} minimized the information loss of the enhanced underwater images to obtain the transmission map and improved the brightness and contrast of underwater images based on natural image histogram distribution prior.
Wang et al.\cite{ref16} proposed the maximum attenuation identification method for UIE based on a simplified underwater light propagation model.
Ancuti et al.\cite{ref4} combined the Laplacian pyramid and white balance to produce two enhanced results, which were then fused via weight mapping.
Peng et al.\cite{ref17} generalized the common DCP method to image restoration. They estimated scene transmission through calculating the difference between the ambient light and the observed intensity. In addition, they incorporated adaptive color correction into the image formation model to restore contrast and remove color casts.
Ma et al.\cite{ref18} proposed a new Retinex variational model with information entropy smoothing and non-uniform illumination priors to enhance underwater images, which models underwater illumination as an independent and piecewise identical distribution to describe the complicated underwater illumination environment and accommodates the traditional Gaussian distribution as a special case.

Although these methods have specific theoretical support,
they may still suffer from under-enhancement or over-enhancement under challenging water environment and lighting condition.
One reason is that the underwater environment is complex and changeable, and physical parameters are difficult to get; the other reason is that it is hard to establish the underwater image degradation model.
\subsection{Deep-Learning-Based UIE Method}
With tremendous progress of deep learning in computer vision, deep-learning-based methods have emerged as the baseline for UIE tasks recently.
The deep learning-based methods trained with different types of underwater images can break through the limitation of conventional methods.
Deep-learning-based methods mainly consist of two categories: CNN-based and GAN-based.

CNN extracts hierarchical image features based on convolution and pooling, enabling automatic learning of effective features in underwater images.
Among them, Sun et al.\cite{ref19} proposed a deep pixel-to-pixel network with an encoder–decoder framework for UIE. It uses convolution layers as encoder for noise filtering, while uses deconvolution layers as decoder for detail recovery and image refinement, achieving self-adaptive data-driven image enhancement without considering the physical environment.
Li et al.\cite{ref20} proposed a gated fusion network called Water-Net for UIE, which fuses the input images into the enhanced image according to the predicted confidence maps. The predicted confidence maps learnt by Water-Net decide the most important features of input images left in the enhanced image.
Li et al.\cite{ref21} presented a light-weight CNN (UWCNN) on the basis of underwater scene prior, which is trained with the corresponding data to accommodate various underwater scenes.
Naik et al.\cite{ref22} proposed a shallow neural network architecture (Shallow-UWnet) for UIE, which has fewer parameters than the state-of-art models while preserving good performance.
Li et al.\cite{ref23} utilized CNN coupled with attention mechanism to extract the most discriminative features from multiple color spaces and adaptively integrate and highlight them, which enhances the quality of underwater images effectively.
These methods above benefit from the efficacy and generalization ability of CNN, which brings considerable improvement to UIE. However, the finite receptive field restricts CNN-based methods from modeling long-range pixel dependence, and the fixed weight of CNN also affects its adaptation to different input contents.

GAN is a generative model based on deep learning. In recent years, GAN has been extensively applied in UIE to improve the usability and visual quality of underwater images.
For example, Li et al.\cite{ref24} proposed WaterGAN to enable real-time color correction for underwater images, which can produce realistic underwater images from in-air image and depth map pairs in an unsupervised manner, and coarsely estimate the depth of underwater scenes.
Islam et al.\cite{ref25} proposed FUnIEGAN for real-time UIE based on conditional GAN. A comprehensive objective function which associates image content, global similarity, and local style and texture is introduced in adversarial training, resulting in better perceptual image quality for both paired and unpaired training.
Liu et al.\cite{ref26} presented MLFcGAN which enhances the contrast and color of underwater images by fusing multi-level features. MLFcGAN augments local features in each level with global features, improving the learning speed and the performance in detail preservation and color correction of the network.
Han et al.\cite{ref27} introduced a light-weight encoder-decoder architecture (UIENet) to enhance underwater images from visual sensors. UIENet is further involved into a generative adversarial network (UIEGAN) model against a supervised discriminator to improve its correction capability for the photorealistic images with more local details and global information.
Cong et al.\cite{ref28} introduced a physical-model-guided GAN for UIE to promote the visual aesthetics and reality of the enhanced images.
Gonzalez et al.\cite{ref29} employed a conditional generative adversarial network with two generators named DGD-cGAN to remove the haze and colour cast induced by the water column and restore the true colours of underwater scenes.
However, the training processes of these GAN-based UIE methods are often divergent or unstable, and the generated images often suffer from uncertainty or a lack of diversity. Therefore, it is difficult to guarantee the consistency and accuracy of generated images.

Although deep learning-based methods have greatly promoted the development of UIE recently, there are still some challenges to overcome.
\subsection{Denoising Diffusion Probabilistic Models}
DDPMs have made great success in image generation, which outperforms GANs, the previous SOTA method, in many tasks.
DDPM is a class of simplified diffusion model which applies variational inference for modeling and reparameterization trick for sampling. In general, the DDPM includes two processes: the forward noising process (i.e., diffusion process) and the reverse denoising process.
The forward noising process gradually injects Gaussian noise to the clean image, making it more and more random and blurry until it approximates an isotropic Gaussian distribution.
The reverse denoising process reconstructs the original image from a Gaussian-distributed noisy image by gradually removing noise.
The reverse process trains a neural network to estimate the conditional probability distribution at each step, more specifically, the data distribution of the previous step given the current data.
DDPM has demonstrated the advantage of more stable training process and higher quality and diversity of synthesis.

Recent advances of DDPMs have been largely due to reformulating DDPMs to estimate noise rather than pixels\cite{ref30}, utilizing cascaded DDPM pipelines in which low-resolution base diffusion models are trained in concurrence with upsamplers\cite{ref31,ref32}, and improved sampling techniques\cite{ref30,ref33,ref34}.
For example, Nicholas et al.\cite{ref14} introduced improved DDPM, which can obtain better log-likelihood and high-quality samples.
Saharia et al.\cite{ref32} presented a class-conditional diffusion model named cascaded diffusion models (CDM) for high-fidelity image generation, which utilizes category labels as input conditions to produce images of the corresponding category.
Saharia et al.\cite{ref10} also proposed super-resolution via repeated refinement (SR3).
SR3 is a conditional DDPM that uses a low-resolution image as input condition and produces the corresponding high-resolution image through an iterative stochastic denoising process.

Currently, DDPMs have been utilized in image reconstruction.
Kawar et al.\cite{ref35} introduced denoising diffusion restoration models (DDRMs) which utilizes a pre-trained denoising diffusion generative model to solve a linear inverse problem since image restoration can usually be represented as a linear inverse problem.
DDRM denoises a sample to the desired output gradually and stochastically, conditioned on the inverse problem and measurements.
In this way a variational inference objective is employed to learn the posterior distribution, from which images are produced efficiently.
Lu et al.\cite{ref36} first presented a DDPM-based UIE method (UW-DDPM), which uses two U-Net networks for image distribution transformation and image denoising, effectively enhancing the underwater images.
However, that work does not address the issue of slow sampling speed of DDPMs and its application scenarios are substantially limited.
Lu et al.\cite{ref37} further proposed SU-DDPM for UIE, which reduces the sampling step and changes the initial sampling distribution to accelerate the sampling inference process and achieve real-time enhancement.
Additionally, SU-DDPM combines the reference image with the degraded image in the diffusion process, effectively solving the issue of color deviation and improving the quality of underwater images.
Song et al.\cite{ref38} introduced a two-stage frequency domain-based latent diffusion model (FD-LDM) for UIE, which firstly employs a lightweight parameter estimation network to estimate the degradation parameters of underwater images, thereby mitigating the impact of color bias on the diffusion model, then extracts high and low-frequency priors and integrates them with the refined latent diffusion model to enhance the images further.
U-Nets\cite{ref12} are presently the prevailing choice for backbone of DDPMs.
Concurrently a new DDPM architecture based on attention mechanism is introduced\cite{ref39}.
Inspired by these diffusion models mentioned above, we propose ICDT to enhance underwater images.
\section{Image-Conditional Diffusion Transformer}
Conditional diffusion models take supplemental information as input.
For UIE task, we use degraded image as the extra conditional input.
Additionally, a few simple modifications are done following\cite{ref14} to make DDPMs achieve better log-likelihoods and sample much faster without sacrificing sample quality.
\subsection{Image-Conditional Denoising Diffusion Probabilistic Models}
DDPMs are a kind of generative models which progressively transforms a Gaussian noise distribution into the distribution of data on which the model is trained. The forward noising process is a Markov process which gradually corrupts the data ${{\bf{x}}_0} \sim q({{\bf{x}}_0})$ by injecting Gaussian noise with variance $\beta _t  \in (0,1)$ at time $t$
\begin{align}
    q({\bf{x}}_{1:T} |{\bf{x}}_0 ) & = \prod\limits_{t = 1}^T q ({\bf{x}}_t |{\bf{x}}_{t - 1} ), \\
    q({{\bf{x}}_t} |{{\bf{x}}_{t - 1}}) & = {\cal N}({{\bf{x}}_t};\sqrt {1 - {\beta _t}} {{\bf{x}}_{t - 1}},{\beta _t}{\bf{I}}).
\end{align}
Given sufficiently large $T$ and an appropriate schedule of $\beta _t$, ${\bf{x}}_T$ approximately obeys an isotropic Gaussian distribution. Thus, if the reverse distribution $q ({\bf{x}}_{t - 1} |{\bf{x}}_t )$ is known, we can start from ${{\bf{x}}_T} \sim {\cal N}(0, {\bf{I}})$ and run the reverse process to obtain a sample from $q({{\bf{x}}_0})$.

\begin{figure}[!t]
\centering
\includegraphics[width=3.49in]{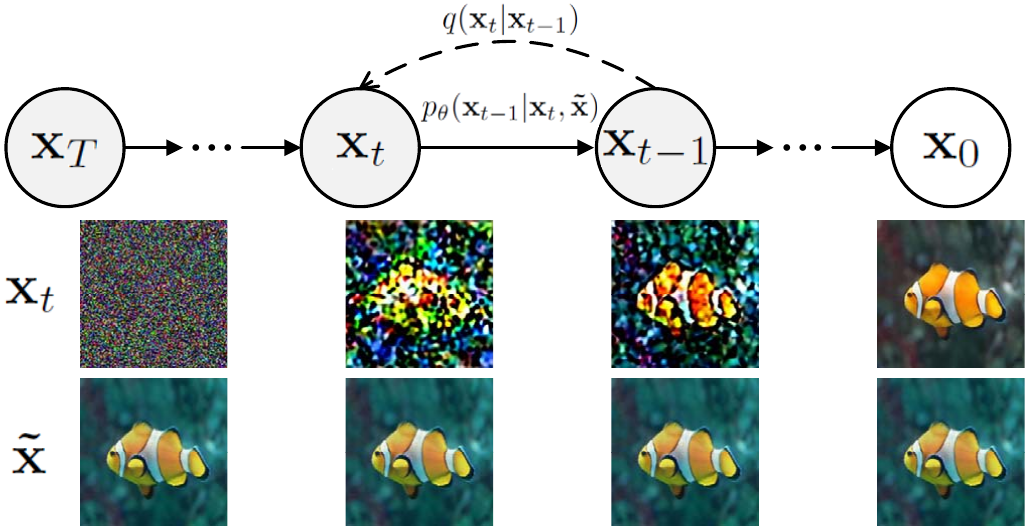}
\caption{The forward noising process (dashed line) and the reverse denoising process (solid line) of an image-conditional DDPM.}
\label{fig_1}
\end{figure}

During training we sample a data pair $({\bf{x}}_0 ,{\bf{\tilde x}})$ where ${\bf{\tilde x}}$ is a degraded image and ${\bf{x}}_0$ is the corresponding ground truth image.
Conditional diffusion models use a deep neural network (DNN) to learn the reverse process defined by the conditional joint distribution
\begin{align}
    p_\theta  ({\bf{x}}_{0:T} |{\bf{\tilde x}}) & = p({\bf{x}}_T )\prod\limits_{t = 1}^T {p_\theta  ({\bf{x}}_{t - 1} |{\bf{x}}_t ,{\bf{\tilde x}})}, \\
    p_\theta  ({\bf{x}}_{t - 1} |{\bf{x}}_t ,{\bf{\tilde x}}) & = {\cal N}({\bf{x}}_{t - 1} ;{\boldsymbol{\mu }} _\theta  ({\bf{x}}_t ,{\bf{\tilde x}},t),{\bf{\Sigma }}_\theta  ({\bf{x}}_t ,{\bf{\tilde x}},t)),\label{eq.4}
\end{align}
so that the sampled image in the diffusion process has high fidelity to the data distribution conditioned on ${\bf{\tilde x}}$ (cf. Fig.~\ref{fig_1}).
Here ${\boldsymbol{\mu }} _\theta  ({\bf{x}}_t ,{\bf{\tilde x}},t)$ and ${\bf{\Sigma }}_\theta  ({\bf{x}}_t ,{\bf{\tilde x}},t)$, the statistics of $p_\theta$, are estimated by the DNN where ${\bf{\tilde x}}$ is provided as input.

We use a variational lower bound (VLB)\cite{ref40} of the negative log likelihood $\mathbb{E}_{q({\bf{x}}_0 )} [ - \log p_\theta  ({\bf{x}}_0 )] \le L_{\rm{vlb}}$ as the loss function which the DNN is trained with. The VLB can be written as\cite{ref30,ref31}
\begin{align}\label{eq.5}
    {L_{\rm{vlb}} } &= \mathbb{E}_q\left[ {\underbrace {{D_{{\rm{KL}}}}(q({{\bf{x}}_T}|{{\bf{x}}_0})||p({{\bf{x}}_T}))}_{{L_T}}\underbrace { - \log {p_\theta }({{\bf{x}}_0}|{{\bf{x}}_1},{\bf{\tilde x}})}_{{L_0}}} \right.  \nonumber\\
    & \hspace{1.1em}\left. { + \sum\limits_{t > 1} {\underbrace {{D_{{\rm{KL}}}}(q({{\bf{x}}_{t - 1}}|{{\bf{x}}_t},{{\bf{x}}_0})||{p_\theta }({{\bf{x}}_{t - 1}}|{{\bf{x}}_t},{\bf{\tilde x}}))}_{{L_{t - 1}}}} } \right].
\end{align}
The loss can be efficiently optimized through stochastic gradient descent over uniformly sampled $L_{t - 1}$ terms\cite{ref30}, taking into account that we can sample intermediate ${\bf{x}}_t$ from an arbitrary step of the diffusion process with the marginal
\begin{equation}\label{eq.6}
    q({{\bf{x}}_t}{\rm{|}}{{\bf{x}}_0}) = {\cal N}({{\bf{x}}_t};\sqrt {{{\bar \alpha }_t}} {{\bf{x}}_0},(1 - {\bar \alpha _t}){\bf{I}}),
\end{equation}
or
\begin{equation}\label{eq.7}
    {{\bf{x}}_t} = \sqrt {{{\bar \alpha }_t}} {{\bf{x}}_0} + \sqrt {1 - {{\bar \alpha }_t}} {{\boldsymbol{\epsilon }}_t},
\end{equation}
where ${\alpha _t} = 1 - {\beta _t}$, ${\bar \alpha _t} = \prod\nolimits_{i = 0}^t {{\alpha _i}}$, and ${\boldsymbol{\epsilon }} _t \sim {\cal N}({\bf{0}},{\bf{I}})$.
The $L_{t - 1}$ term in (\ref{eq.5}) is a KL divergence between two Gaussian distributions, $p_\theta  ({\bf{x}}_{t - 1} |{\bf{x}}_t ,{\bf{\tilde x}})$ from (\ref{eq.4}) and $q({{\bf{x}}_{t - 1}}|{{\bf{x}}_t},{{\bf{x}}_0})$.
The latter is the generative process posterior, which can be calculated using Bayes theorem
\begin{equation}\label{eq.8}
    q({{\bf{x}}_{t - 1}}{\rm{|}}{{\bf{x}}_t},{{\bf{x}}_0}) = {\cal N}({{\bf{x}}_{t - 1}};{{\boldsymbol{\tilde \mu }} _t}({{\bf{x}}_t},{{\bf{x}}_0}),{\tilde \beta _t}{\bf{I}})
\end{equation}
where the parameters of the distribution are defined as
\begin{align}
    {{\boldsymbol{\tilde \mu }} _t}({{\bf{x}}_t},{{\bf{x}}_0}) & = \frac{{\sqrt {{{\bar \alpha }_{t - 1}}} {\beta _t}}}{{1 - {{\bar \alpha }_t}}}{{\bf{x}}_0} + \frac{{\sqrt {{\alpha _t}} (1 - {{\bar \alpha }_{t - 1}})}}{{1 - {{\bar \alpha }_t}}}{{\bf{x}}_t},\label{eq.9} \\
    {\tilde \beta _t} & = \frac{{(1 - {{\bar \alpha }_{t - 1}})}}{{(1 - {{\bar \alpha }_t})}}{\beta _t}.
\end{align}

Since both $q$ and $p_\theta$ are Gaussian, we evaluate the KL divergence with the mean and covariance of them. The mean ${\boldsymbol{\mu }} _\theta  ({\bf{x}}_t ,{\bf{\tilde x}},t)$ can be parameterized in many different ways. For example, we could predict ${\boldsymbol{\mu }} _\theta  ({\bf{x}}_t ,{\bf{\tilde x}},t)$ directly or predict ${\bf{x}}_0$ and use it in (\ref{eq.9}) to get ${\boldsymbol{\mu }} _\theta  ({\bf{x}}_t ,{\bf{\tilde x}},t)$.
Alternatively, one can predict the noise ${\boldsymbol{\epsilon }}_t$ and incorporate (\ref{eq.7}) into (\ref{eq.9}) to derive\cite{ref23}
\begin{equation}\label{eq.11}
    {\boldsymbol{\mu }} _\theta  ({\bf{x}}_t ,{\bf{\tilde x}},t) = \frac{1}{{\sqrt {{\alpha _t}} }}\left( {{{\bf{x}}_t} - \frac{{{\beta _t}}}{{\sqrt {1 - {{\bar \alpha }_t}} }}{{\boldsymbol{\epsilon }}_\theta }({\bf{x}}_t ,{\bf{\tilde x}},t)} \right).
\end{equation}
For image-conditional generation, ${\bf{x}}_t$ and ${\bf{\tilde x}}$ are channel-wise concatenated as the input of the DNN.
In this setting the DNN is trained to predict ${\boldsymbol{\epsilon }}_t$ by optimizing the  mean-squared error objective
\begin{equation}\label{eq.12}
    {L_{{\rm{simple}}}} = {\mathbb{E}_{{{\bf{x}}_0},t,{\boldsymbol{\epsilon }}_t}}[||{{\boldsymbol{\epsilon }}_t} - {{\boldsymbol{\epsilon }}_\theta }({{\bf{x}}_t},{\bf{\tilde x}},t)|{|^2}].
\end{equation}
This simplified objective can be viewed as a re-weighted version of $L_{\rm{vlb}}$, which omits the terms involving ${\bf{\Sigma }}_\theta$.

To improve the log-likelihood of ${\bf{x}}_0$, a better choice of ${\bf{\Sigma }}_\theta  ({\bf{x}}_t ,{\bf{\tilde x}},t)$ is needed. Since $L_{{\rm{simple}}}$ provides no learning guidance for ${\bf{\Sigma }}_\theta  ({\bf{x}}_t ,{\bf{\tilde x}},t)$, a hybrid objective is introduced\cite{ref14}
\begin{equation}\label{eq.13}
    {L_{{\rm{hybrid}}}} = {L_{{\rm{simple}}}} + \lambda {L_{{\rm{vlb}}}},
\end{equation}
where ${\bf{\Sigma }}_\theta  ({\bf{x}}_t ,{\bf{\tilde x}},t)$ is learnt via $L_{{\rm{vlb}}}$ while $L_{{\rm{simple}}}$ remains the main source of impact on ${\boldsymbol{\mu }} _\theta  ({\bf{x}}_t ,{\bf{\tilde x}},t)$ with small $\lambda$ (e.g., $\lambda = 0.001$) to keep $L_{{\rm{vlb}}}$ from overwhelming $L_{{\rm{simple}}}$.

Additionally, it has been shown that incorporating learnt variances into DDPMs allows sampling with much fewer diffusion steps than training with a negligible change in sample quality\cite{ref14}.
During sampling we only need a sub-sequence of the complete $\{1,...,T\}$ timestep indices, e.g. a sub-sequence with uniformly spaced integers between $1$ and $T$.
While DDPM needs hundreds of sampling steps to generate high-quality images, we can achieve almost the same quality with an order of magnitude fewer sampling steps, thus accelerating sampling for deployment in practical applications.
\subsection{Architecture of Image-Conditional Diffusion Transformer}
\begin{figure*}[!t]
\centering
\includegraphics[width=4.6in]{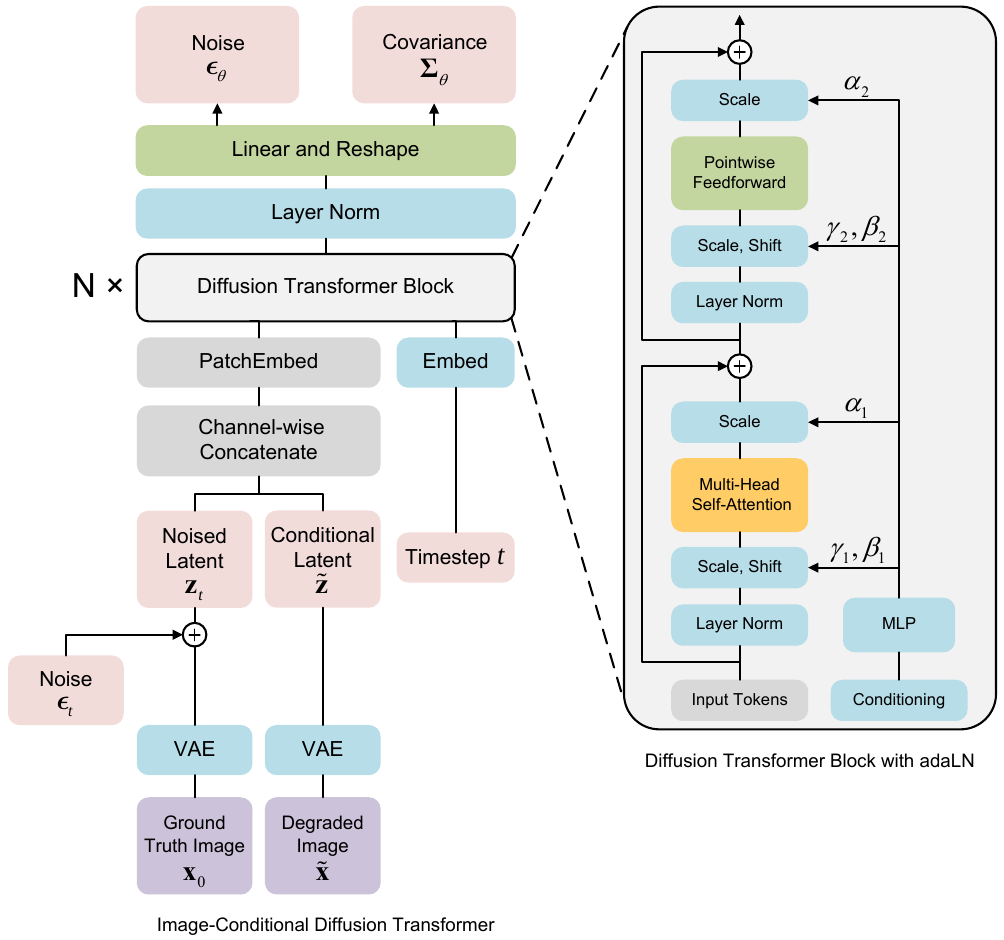}
\caption{The ICDT architecture. Left: We train ICDT models in latent space. The converted latent is divided into patches, which subsequently pass through $N$ diffusion transformer blocks. Right: Diffusion transformer blocks which incorporate conditioning through adaptive layer norm.}
\label{fig_2}
\end{figure*}

The architecture of the proposed ICDT is based on diffusion transformer.
Since the focus of diffusion transformer is on learning spatial representations of images, it operates on sequences of patches like vision transformer (ViT)\cite{ref13}. Moreover, diffusion transformer retains many good practices of ViTs. Fig.~\ref{fig_2} illustrates the ICDT architecture and its forward pass will be described in detail below.

It is computationally prohibitive to train diffusion models in pixel space directly, especially for large-size images. Latent diffusion models (LDMs)\cite{ref41} address the problem in two steps: (1) train an autoencoder which compresses images into small spatial representations via a learned encoder ${\bf{z}} = E(\bf{x})$; (2) train a diffusion model for representations $\bf{z}$ rather than for images $\bf{x}$ ($E$ is fixed). Generated images can be obtained by decoding a sample $\bf{z}$ from the diffusion model to an image via the learned decoder ${\bf{x}} = D(\bf{z})$.
LDMs achieve comparable performance while occupying much fewer FLOPs than pixel space diffusion models, which makes them an attractive option for architecture design.
In consideration of computational efficiency, we employ diffusion transformers to latent space, which renders the image generation pipeline a hybrid architecture. Specifically, in the hybrid model we apply pre-trained VAEs and transformer-based DDPMs.

For image-conditional diffusion models, the conditional image (degraded image in this setting) $\bf{\tilde x}$ is provided as input.
As shown in Fig.~\ref{fig_2}, the input images ${\bf{x}}_0$ and $\bf{\tilde x}$ pass through the VAEs, converted to latents ${\bf{z}}_0$ and $\bf{\tilde z}$ with shape $I \times I \times C$.
Then, the noised latent ${\bf{z}}_t$ is obtained according to (\ref{eq.7}). We concatenate ${\bf{z}}_t$ and $\bf{\tilde z}$ channel-wise, doubling the input channel dimension for diffusion transformer.

The first layer of ICDT is “patchembed”, which implements patchifying and embedding. Patchifying reshapes the spatial representations ${\bf{z}}_t$ and $\bf{\tilde z}$ into a sequence of $T$ flattened 2D patches.
The number of patches $T$ depends on the patch size hyperparameter $p$, more specifically, $T=(I/p)^2$. Doubling $P$ will quarter $T$, and thus quarter total transformer FLOPs at least. Although it significantly affects FLOPs, changing $p$ does not substantially change the parameter-counts of downstream modules.
Embedding maps each flattened 2D patch to a $d$-dimensional token via a trainable linear projection, where $d$ is the constant hidden dimension size used by the transformer through all layers.
Then, we add standard ViT frequency-based positional embeddings to all tokens to retain positional information.

Following patchembed, the resulting tokens serve as an input to $N$ successive diffusion transformer blocks, each of which operates at hidden size $d$. Another input is the embedding vector of the noise timestep $t$.
We utilize a 256-dimensional frequency embedding\cite{ref31} followed by a two-layer MLP with dimension equal to the transformer’s hidden size and SiLU activations to embed the noise timestep.
We apply a variant of transformer block named adaptive layer norm (adaLN) block, which is the most compute-efficient of the three block designs explored in\cite{ref42}.
As shown in Fig.~\ref{fig_2}, adaLN block replaces standard layer norm layers in ViT. We regress shift parameters $\boldsymbol{\beta}$ and dimension-wise scale parameters $\boldsymbol{\alpha}$ and $\boldsymbol{\gamma}$ from the timestep embedding instead of learning them directly.
We adopt standard transformer configurations which jointly scale $N$, $d$ and attention heads, mimicking ViT\cite{ref13,ref43}.
Specifically, we adopt four configurations: ICDT-S, ICDT-B, ICDT-L, and ICDT-XL, which are detailed in Table \ref{tab:table1}.
They encompass a broad spectrum of FLOP allocations and model sizes, ranging from 0.36 to 118.64 GFLOPs, enabling us to measure scalability properties.

\begin{table*}[!t]
\caption{Details of ICDT Models\label{tab:table1}}
\centering
\begin{tabular}{l c c c c c}
\toprule[1pt]
Model   &  Heads  &  Hidden size $d$  &  Layers $N$  &  Params (M)  &  FLOPs (G) \\
\midrule
ICDT-S   &  6        &  384            &  12      &  32.97      &  1.41       \\
ICDT-B   &  12        &  768            &  12     &  130.52     &  5.56       \\
ICDT-L   &  16        &  1024           &  24     &  458.12     &  19.70      \\
ICDT-XL  &  16        &  1152           &  28     &  675.15     &  29.05      \\
\bottomrule[1pt]
\end{tabular}
\begin{tabular}{l}
\toprule[0pt]
The values of Params (M) and FLOPs (G) are corresponding to $I =$ 32 \\
and $p =$ 4.
\end{tabular}
\end{table*}

Following the last diffusion transformer block, the sequence of tokens is decoded into a diagonal covariance prediction and a noise prediction, which both have the same shape as the input latent ${\bf{z}}_t$ (or $\bf{\tilde z})$.
To accomplish this, we apply a standard linear decoder.
Specifically, we use the final adaptive layer norm and linearly decode each token into a $p \times p \times 2C$ tensor, where $C$ is the channel dimension of the latent input to diffusion transformer blocks.
Eventually, according to their original spatial layout we rearrange the decoded tensors to obtain the covariance and noise predictions.
\section{Experimental Setup}
We first explore the scaling properties of our ICDT models. In this paper, most analysis of the model complexity is from the perspective of FLOPs which is widely used in the architecture design literature.
We gauge the scalability of our ICDT models from the standpoint of forward pass complexity measured by FLOPs.
The models are briefly notated with their configurations and input latent patch sizes $p$ in the following; for instance, ICDT-XL/2 means the XLarge configuration and $p =$ 2. We provide three choices for $p$ (i.e., $p =$ 2, 4, 8).
Choosing smaller $p$ leads to longer sequence length $T$ and thus more FLOPs.
\subsection{Datasets}
We use the Underwater ImageNet\cite{ref8} dataset, a subset of the enhancement of underwater visual perception (EUVP)\cite{ref25} dataset.
This dataset covers a wide scope of scene/main object categories such as coral and marine life.
The EUVP dataset contains separate sets of unpaired and paired image samples of poor and good perceptual quality to facilitate supervised training of UIE models, which also offers a platform to evaluate the performance of different UIE methods.
The Underwater ImageNet dataset includes 3700 paired underwater images.
We randomly select 3328 pairs of the images from it to create the training set, while the remaining 372 pairs constitute the test set.
\subsection{Diffusion Model}
We use off-the-shelf pre-trained VAEs (ft-EMA)\cite{ref40} from Stable Diffusion \cite{ref41}. The VAE encoder $E(\cdot)$ has a downsample factor of 8—for an image $\bf{x}$ with shape 256 $\times$ 256 $\times$ 3, the output latent $\bf{z}$ has shape 32 $\times$ 32 $\times$ 4.
All of our ICDT models work in this latent space. After sampling a latent $\bf{z}$ from the diffusion model, we convert it into pixel space via the VAE decoder ${\bf{x}} = D(\bf{z})$.
As for diffusion hyperparameters, (1) we adopt a $T =$ 1000 linear schedule of $\beta _t$ spanning from 1 $\times$ 10$^{ - 4}$ to 2 $\times$ 10$^{ - 2}$; (2) we parameterize the covariance ${\bf{\Sigma }}_\theta$ according to (15) in \cite{ref14} based on which the learning of ${\bf{\Sigma }}_\theta$ works; (3) we use a 256-dimensional frequency embedding \cite{ref31} succeeded by a two-layer multilayer perceptron (MLP) to embed the timestep $t$.

We used the same configurations and hyperparameters for all of our ICDT models. Further specifications of our models are provided in Table III of Supplementary Materials, available online.
\subsection{Training}
We train our diffusion models at image resolution of 256 $\times$ 256 on the Underwater ImageNet dataset.
All images in the training set were used per epoch.
We do not use any data augmentation other than horizontal flips.
All models are trained with a batch size of 32 and optimized by AdamW\cite{ref44,ref45} with a fixed learning rate of 1 $\times$ 10$^{ - 4}$ and no weight decay.
We do not apply learning rate warmup or regularization which are commonly used in ViTs\cite{ref46,ref47} since we find them unnecessary to train our diffusion models to high performance stably.
In accordance with common practice when training generative models, we retain an exponential moving average (EMA) with a decay of 0.9999 during model parameter updates to make learning more stable\cite{ref43,ref48}.

Previous research on ResNets has revealed the benefits of initializing a residual block to be the identity function. Here we perform the same initialization for the adaLN diffusion transformer block.
Considering the dimension-wise scale parameters $\alpha$ applied immediately before each residual connections in each diffusion transformer blocks, the MLP is initialized to output $\boldsymbol{\alpha}$ as zero vector, making the whole diffusion transformer block to be the identity function.
In addition, we also initialize the final adaptive linear layer with zeros.

We employ the same training hyperparameters and initialization for all model configurations and latent patch sizes.
All models are trained on a single NVIDIA RTX A6000 GPU. ICDT-XL/2, the most computational heavy model, is trained at approximately 0.59 iteration per second.
\subsection{Evaluation Metrics}
We assess scaling properties of our models and compare them with prior works quantitatively. The evaluation metrics include peak signal-to-noise ratio (PSNR)\cite{ref49}, structural similarity (SSIM)\cite{ref50}, learned perceptual image patch similarity (LPIPS)\cite{ref51}, and underwater image quality measure (UIQM)\cite{ref52}. We report these metric scores on the enhanced images generated with 250 DDPM sampling steps.

Since all images in the Underwater ImageNet dataset have corresponding reference images, three full-reference image quality metrics PSNR, SSIM, and LPIPS are employed for quantitative evaluations between enhanced images and reference images.
A higher PSNR or SSIM score denotes the result is closer to the content of the reference image, while A lower LPIPS score indicates higher human perceptual similarity between the enhanced and the reference images.
We present the average PSNR, SSIM, and LPIPS scores of our models and comparison models on the 372 images in the test set.

We also utilize the metric UIQM for non-reference quality assessment of underwater image enhancement performance.
Better perceptual image quality leads to higher UIQM score.
Average UIQM scores of enhanced images generated from the 372 images in the test set with our models and comparison methods are presented as well.
\section{Experiments}
\subsection{Scalability Properties}
\begin{figure*}[!t]
\centering
\includegraphics[width=6.5in]{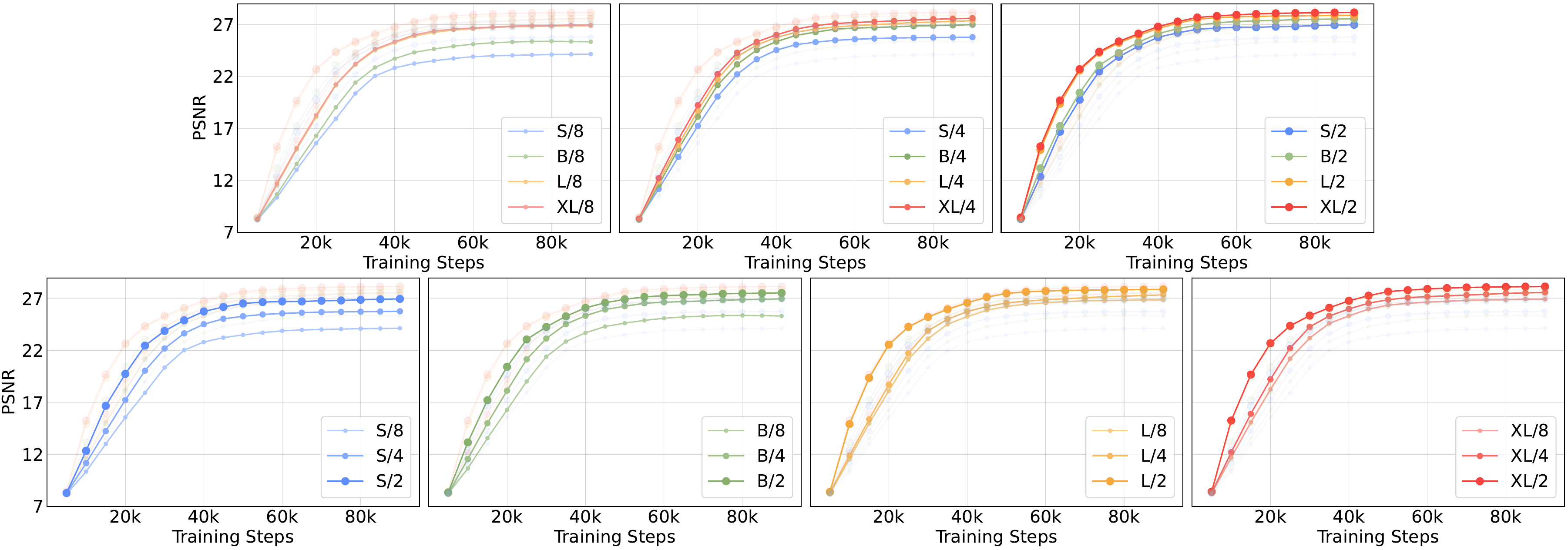}
\caption{Scaling the ICDT model improves PSNR during the whole training process. We present PSNR over training steps for all 12 ICDT variants. In the top row, we compare PSNR against model size while keeping patch size the same. In the bottom row, we compare PSNR against patch size while keeping model size the same.}
\label{fig_3}
\end{figure*}

We train 12 models covering four model configurations (XL, L, B, S) and three different choices of patch size (2, 4, 8).
Three top subfigures in Fig.~\ref{fig_3} show how PSNR changes with model size when patch size is kept unchanged.
Throughout all model configurations, considerable improvements of PSNR are attained during the whole training process by using deeper and wider transformer.
Similarly, four bottom subfigures in Fig.~\ref{fig_3} illustrate PSNR when patch size is changing while model size is fixed.
Significant PSNR improvements are again observed across training through merely increasing the number of tokens input to the models, holding parameters roughly constant.
In summary, we discover that decreasing patch size and increasing model size can substantially improve the image enhancement performance of our diffusion models.

\begin{figure}[!t]
\centering
\includegraphics[width=3.49in]{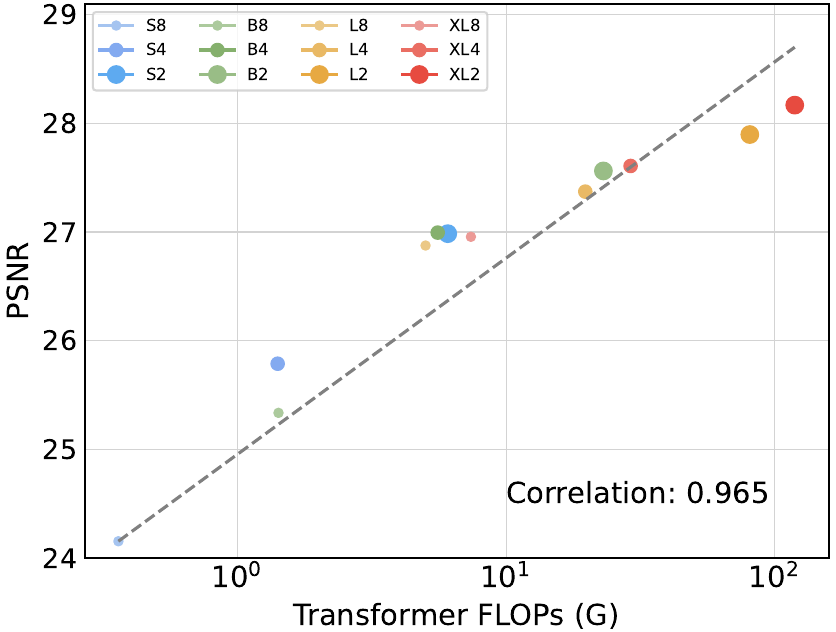}
\caption{PSNR is strongly correlated with model FLOPs. We show PSNR of each ICDT model after 90K training iterations and each model’s FLOPs.}
\label{fig_4}
\end{figure}

The results in Fig.~\ref{fig_3} indicate that parameter-counts is not the unique factor which determines the image enhancement quality of our models. As model size remains unchanged and patch size decreases, the overall amount of the transformer’s parameters is almost constant (slightly decreased actually), while only FLOPs increases. The results suggest that scaling model FLOPs is crucial to the performance of our models.
To clarify this further, we plot the PSNR versus model FLOPs in Fig.~\ref{fig_4}. The results illustrate that different models have comparable performance in terms of PSNR if their total FLOPs are similar (e.g., ICDT-B/4 and ICDT-S/2).
We observe a significant positive correlation between PSNR and model FLOPs, indicating that increased model computation is the key ingredient to improve the models' performance.
We observe the same trend in Fig.~8 of Supplementary Materials for other metrics such as SSIM.

\begin{figure}[!t]
\centering
\includegraphics[width=3.49in]{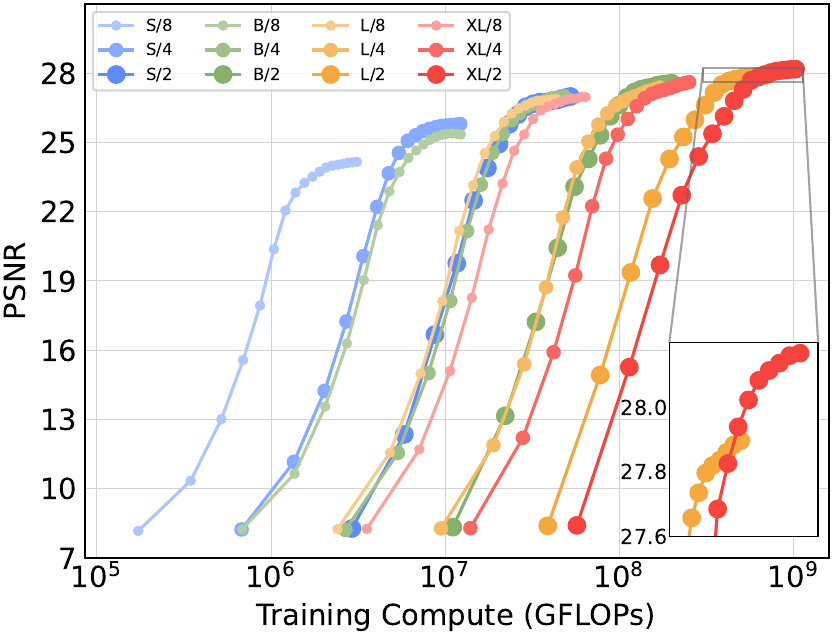}
\caption{Larger ICDT models are more compute-efficient. We present PSNR against total training compute.}
\label{fig_5}
\end{figure}

In Fig.~\ref{fig_5}, we plot PSNR against total training compute for all models, where training compute is approximately estimated as model FLOPs · batch size · training iterations · 3 (the factor of 3 approximates the backwards pass to be twice as compute-intensive as the forward pass).
We find that models with small FLOPs (small model size or large patch size), despite being trained longer, eventually have inferior performance relative to models with large FLOPs trained for fewer iterations.
For example, XL/2 outperforms L/2 after about 6.6 $\times$ 10$^{ 8}$ GFLOPs.
After sufficient training, computational intensive models behavior better even when controlling for total training FLOPs; that is to say, they are more compute-efficient.

\begin{figure*}[!ht]
\centering
\includegraphics[width=6.47in]{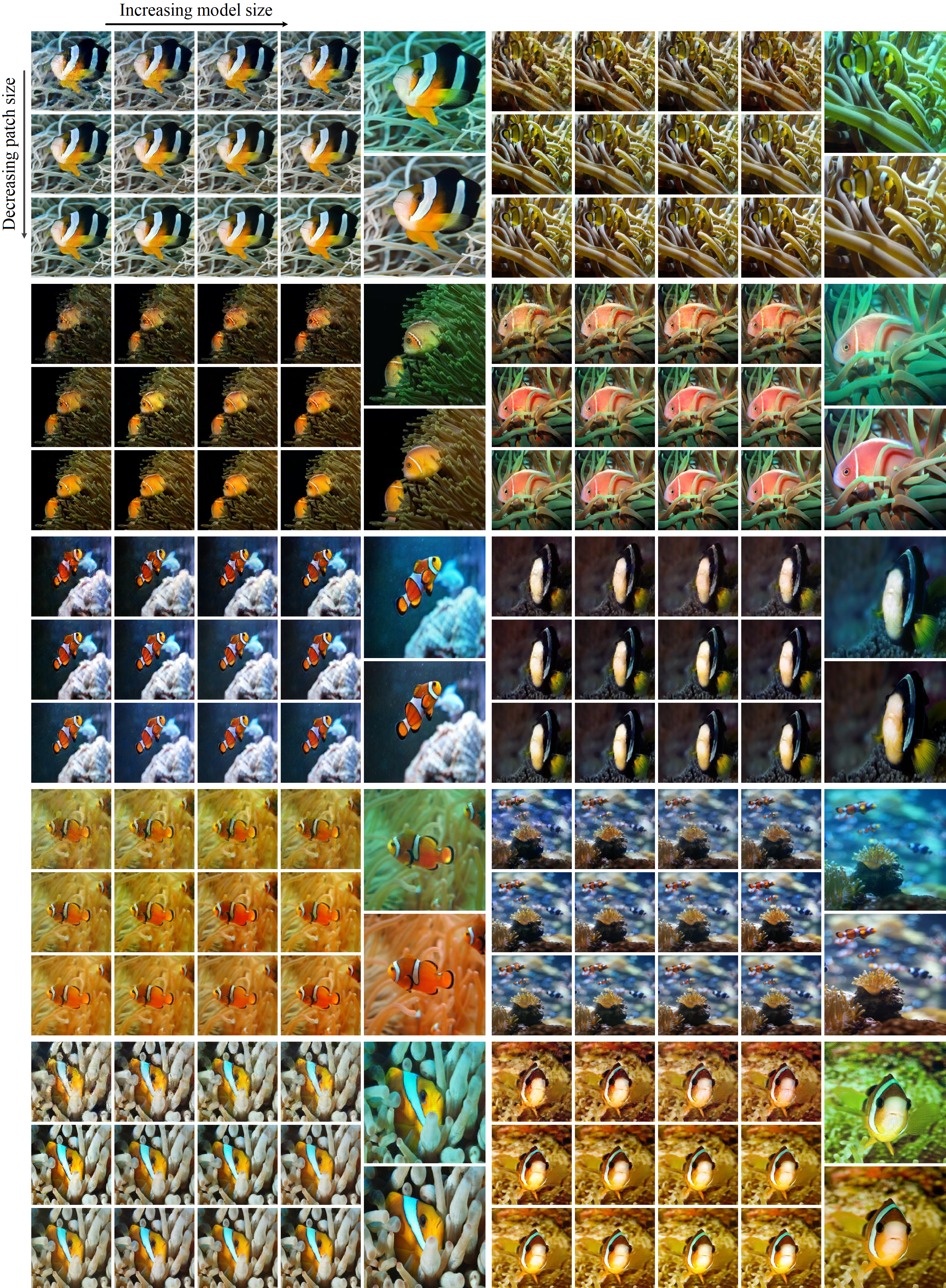}
\caption{Increasing forward pass FLOPs of ICDT improves sample quality. \textit{Best viewed zoomed-in}. Starting from the same input latent noise, we sample from all 12 of our ICDT models after 90K training iterations. The left 12 smaller images in a group (The 14 images containing the same content make up a group.) are the samples from each models, while the two enlarged images in the upper and lower right corners are the degraded image and the reference image, respectively.
Increasing transformer depth/width or the number of input tokens, which equally increases model FLOPs, achieves substantial improvements in visual quality.}
\label{fig_6}
\end{figure*}

We illustrate the impact of scaling on sample quality in Fig.~\ref{fig_6}.
Using the same initial noise ${\bf{x}}_T$ and sampling noise, we sample images from each of the 12 models at 90K training iterations.
This visually demonstrates that increasing either the number of tokens or model size results in considerable improvements in sample quality.
\subsection{Comparison with Other UIE Methods}
After analyzing the scalability, we compare the most computational intensive model, ICDT-XL/2, with SOTA UIE methods.
These methods includes five CNN-based methods (Water-Net\cite{ref20}, UWCNN\cite{ref21}, Shallow-UWnet\cite{ref22}, UIECˆ2-Net\cite{ref53}, and RAUNE-Net\cite{ref54}) as well as two GAN-based methods (FUnIEGAN\cite{ref25} and MLFcGAN\cite{ref26}).

\begin{figure*}[!ht] 
\centering
\includegraphics[width=5.8in]{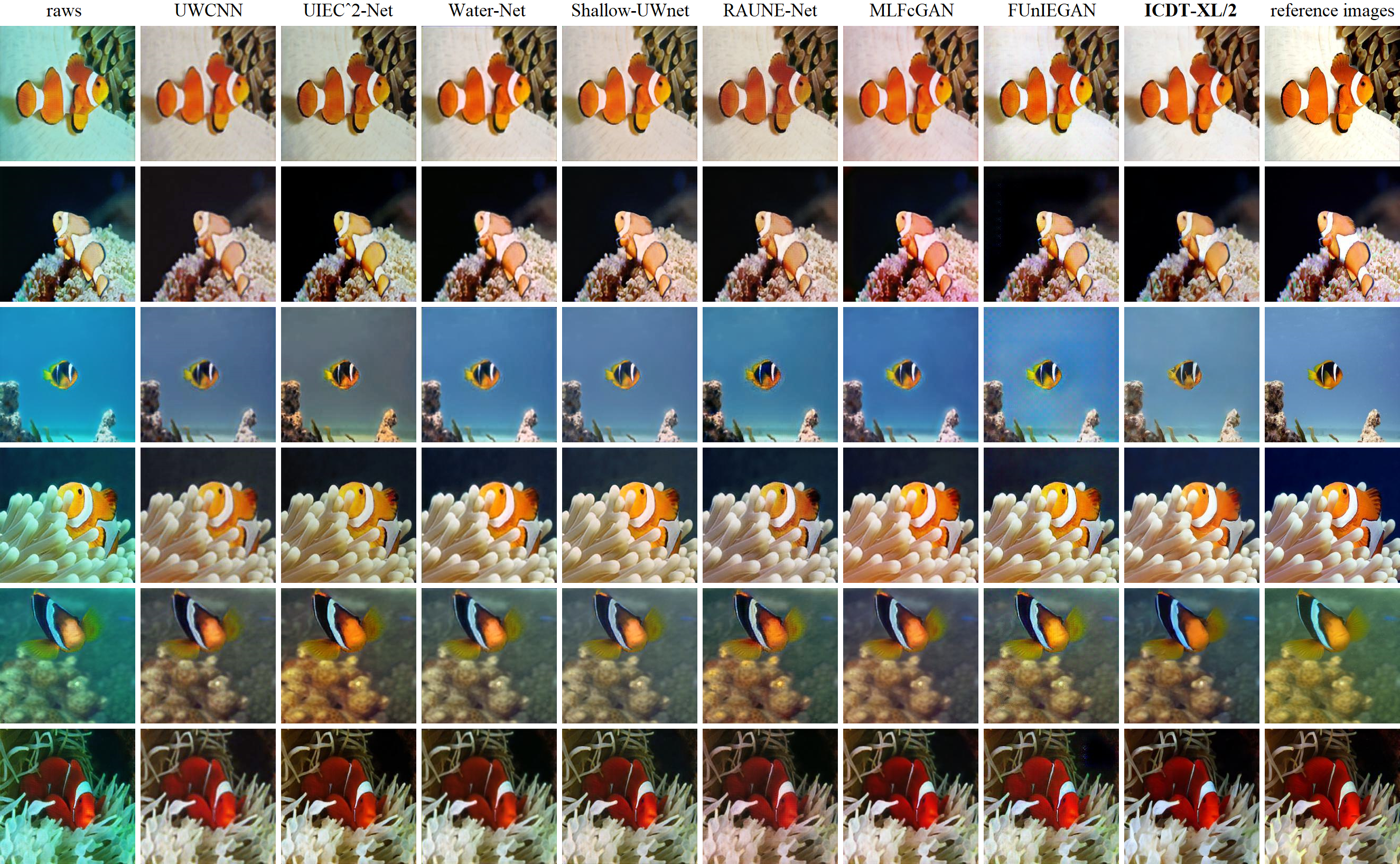}
\caption{Qualitative comparisons of seven different UIE methods and our ICDT method. From left to right are raw images, the results of UWCNN\cite{ref21}, UIECˆ2-Net\cite{ref53}, Water-Net\cite{ref20}, Shallow-UWnet\cite{ref22}, RAUNE-Net\cite{ref54}, MLFcGAN\cite{ref26}, FUnIEGAN\cite{ref25}, the proposed ICDT-XL2, and reference images.}
\label{fig_7}
\end{figure*}

We visualize several results in Fig.~\ref{fig_7}.
In Fig.~\ref{fig_7}, the proposed ICDT effectively eliminates the haze on the underwater images and remits color casts, producing visually pleasing images.
By contrast, the competing methods induce low brightness and contrast (e.g., UWCNN\cite{ref21}, UIECˆ2-Net\cite{ref53}, and FUnIEGAN\cite{ref25}) or introduce blurring (e.g., UWCNN\cite{ref21}), unexpected colors (e.g., MLFcGAN\cite{ref26}), and artifacts (e.g., RAUNE-Net\cite{ref54} and FUnIEGAN\cite{ref25}).

Table~\ref{tab:table2} reports the results of quantitative comparisons with SOTA UIE methods in terms of PSNR, SSIM, LPIPS, and UIQM.
The full-reference image quality metrics PSNR, SSIM, and LPIPS are obtained through comparison between the result of each method and the corresponding reference image.
DiT-XL/2 outperforms all competing methods, achieving the highest PSNR and SSIM as well as the lowest LPIPS.
In addition, we observe that our DiT-XL/2 achieves the highest UIQM score of all prior UIE methods, indicating DiT-XL/2 also performs best in non-reference image quality assessment.
\begin{table}[!t]
\caption{Quantitative Comparisons with SOTA UIE Methods\label{tab:table2}}
\centering
\begin{tabular}{l c c c c}
\toprule[1pt]
Method   &  PSNR$\uparrow$  &  SSIM$\uparrow$  &  LPIPS$\downarrow$  &  UIQM$\uparrow$ \\
\midrule
UWCNN\cite{ref21}   &  17.7347        &  0.7112            &  0.3911      &  2.6407       \\
UIECˆ2-Net\cite{ref53}   &  22.4930        &  0.8254            &  0.2932     &  2.7152       \\
Water-Net\cite{ref20}  &  25.4032        &  0.8355           &  0.2755     &  2.8481      \\
Shallow-UWnet\cite{ref22}   &  26.1325        &  \underline{0.8562}           &  0.2359     &  2.8331      \\
RAUNE-Net\cite{ref54}  &  \underline{26.6946}        &  0.8415           &  0.1990     &  \underline{2.8869}      \\
\midrule
MLFcGAN\cite{ref26}  &  21.6062        &  0.8246           &  0.3646     &  2.7195      \\
FUnIEGAN\cite{ref25}   &  24.1463        &  0.8233           &  \underline{0.1952}     &  2.8239      \\
\midrule
\textbf{ICDT-XL/2}  &  \textbf{28.1673}        &  \textbf{0.8744}           &  \textbf{0.1362}     &  \textbf{2.9234}      \\
\bottomrule[1pt]
\end{tabular}
\begin{tabular}{l}
\toprule[0pt]
Best and second best values are indicated with bold text and underlined      \\
text respectively.
\end{tabular}
\end{table}

Quantitative and qualitative comparisons demonstrate the advantage of our ICDT method.

\section{Conclusion}
We introduce a novel ICDT model for UIE, which uses the degraded underwater image as the extra conditional input. ICDT employs a transformer backbone instead of the U-Net which is commonly used in DDPMs.
As a result, ICDT inherits good scalability properties from the transformer model, enabling a trade-off between image enhancement quality and model complexity.
Through comparisons against prior works, our largest ICDT model achieves SOTA image enhancement quality.
Given the promising experimental results, we will continue to scale ICDTs to larger token-count and  model size in the future.
Note that ICDT is a generic image-to-image generative model. With this understanding, this article can be regarded as taking UIE task as an example to interpret the working of ICDT.
We will explore the application of ICDT in other image-to-image generation tasks such as image translation, image coloring, image super-resolution, image restoration, and image inpainting in the future.

\section*{Acknowledgments}
This work was supported by the Zhenjiang Jinshan Talent Program and the National Natural Science Foundation of China under Grant 52071164.

\vspace{11pt}

\vfill

\end{document}